\documentclass{article} % For LaTeX2e
\usepackage{iclr2027_conference,times}

\usepackage{amsmath,amsfonts,bm}

\def\eqref#1{equation~\ref{#1}}
\def\1{\bm{1}}

\DeclareMathAlphabet{\mathsfit}{\encodingdefault}{\sfdefault}{m}{sl}
\SetMathAlphabet{\mathsfit}{bold}{\encodingdefault}{\sfdefault}{bx}{n}

\usepackage{hyperref}
\usepackage{url}
\usepackage{booktabs}
\usepackage{wrapfig}
\usepackage{multirow}
\usepackage{float}
\usepackage[pdftex]{graphicx}

\usepackage[T1]{fontenc}

\usepackage[most]{tcolorbox}
\usepackage{listings}
\usepackage{needspace}
\definecolor{promptframe}{HTML}{B8C0CC}
\definecolor{promptback}{HTML}{F7F8FA}
\definecolor{prompttitle}{HTML}{E9EDF2}
\definecolor{promptmeta}{HTML}{596273}
\lstdefinestyle{promptstyle}{
    basicstyle=\ttfamily\scriptsize,
    breaklines=true,
    breakatwhitespace=false,
    columns=fullflexible,
    keepspaces=true,
    showstringspaces=false,
    upquote=true,
    tabsize=2,
    aboveskip=0pt,
    belowskip=0pt
}
\newtcblisting{promptbox}[1]{
    enhanced,
    listing only,
    listing engine=listings,
    listing options={style=promptstyle},
    colback=promptback,
    colframe=promptframe,
    colbacktitle=prompttitle,
    coltitle=black,
    fonttitle=\bfseries\small,
    title={#1},
    boxrule=0.45pt,
    arc=1mm,
    left=1.5mm,
    right=1.5mm,
    top=1mm,
    bottom=1mm,
    before skip=5pt,
    after skip=9pt
}
\newcommand{\promptusage}[1]{%
    \noindent{\footnotesize\color{promptmeta}\textit{Usage.} #1}\par\vspace{2pt}}
\newcommand{\promptsubsubsection}[1]{%
    \Needspace{10\baselineskip}\subsubsection{#1}}
\newcommand{\promptagent}[1]{%
    \Needspace{12\baselineskip}\paragraph{#1}}
    
\newcommand{\acc}[2]{#1{\scriptstyle\pm #2}}
\newcommand{\accbold}[2]{\mathbf{#1}{\scriptstyle\pm #2}}

\title{Beyond Answer Sharing: TrueMAD for Targeted Peer Engagement in Multi-Agent Debate}

\author{Dezheng Bao, Sihang Du, Renhong Huang, Zhizhang Yuan, Yifei Sun \& Yang Yang\\
Zhejiang University\\
\texttt{\{baodezheng, yangya\}@zju.edu.cn} \\
}

\iclrfinalcopy % Uncomment for camera-ready version, but NOT for submission.
\begin{document}

\maketitle

\begin{abstract}
    Multi-agent debate (MAD) aims to improve large language model reasoning by
    exposing agents to alternative solutions and enabling iterative revision.
    However, standard broadcast-style MAD makes peer answers visible without
    requiring agents to engage with a particular peer claim. As a result, reliable reasoning may be suppressed by conflicting or majority responses, limiting achievable performance, while erroneous reasoning may persist and degrade final predictions without actionable feedback. We propose \textsc{TrueMAD}, a backbone-agnostic test-time plug-in that introduces
    one-to-one, free-form peer advice between answer broadcasting and
    self-revision. Each source agent responds to a target's current reasoning
    through support, rebuttal, refinement, or further questions, while the target
    retains control over how to revise. The same mechanism can augment standard
    MAD and its improved variants without parameter updates. Across AIME,
    GPQA-Diamond, and TruthfulQA with \texttt{Qwen3-30B-A3B}, \textsc{TrueMAD}
    improves agent-level and strict-majority accuracy over MAD by $3.62$ and
    $3.52$ percentage points on average, while \textsc{TrueDMAD} improves DMAD by
    $4.72$ and $4.06$ points, respectively. Analyses at the utterance,
    agent-transition, and group levels further show more peer-specific
    deliberation, more correction-oriented revisions, and more favorable consensus
    dynamics. 
    % Our code is available at \url{https://anonymous.4open.science/r/TrueMAD-ICLR/}.
\end{abstract}

\section{Introduction}
\label{section-intro}

Large language models (LLMs) are increasingly used for content generation, reasoning, and decision support across domains~\citep{brown2020language,wei2022chain,huang2024large}. However, a single LLM's perspective can limit exploration of alternative paths and recovery from early mistakes on complex tasks~\citep{kojima2022large,huang2024large}. This limitation has motivated multi-agent systems, in which LLMs provide complementary perspectives and feedback. Inspired by the ``society of mind'' view of intelligence~\citep{minsky1986society,zhuge2023mindstorms}, multi-agent debate (MAD) lets agents independently solve problems, exchange responses, and revise answers over rounds. MAD has been explored in mathematical and strategic reasoning, factuality improvement, LLM-based evaluation, and applications~\citep{du2024improving,liang2024encouraging,chan2024chateval,bao2025pd3}. Rather than relying on a single reasoning path, its central advantage is to turn diverse individual perspectives into more reliable individual and collective decisions.

\begin{figure}[t]
  \centering
  \includegraphics[width=\linewidth]{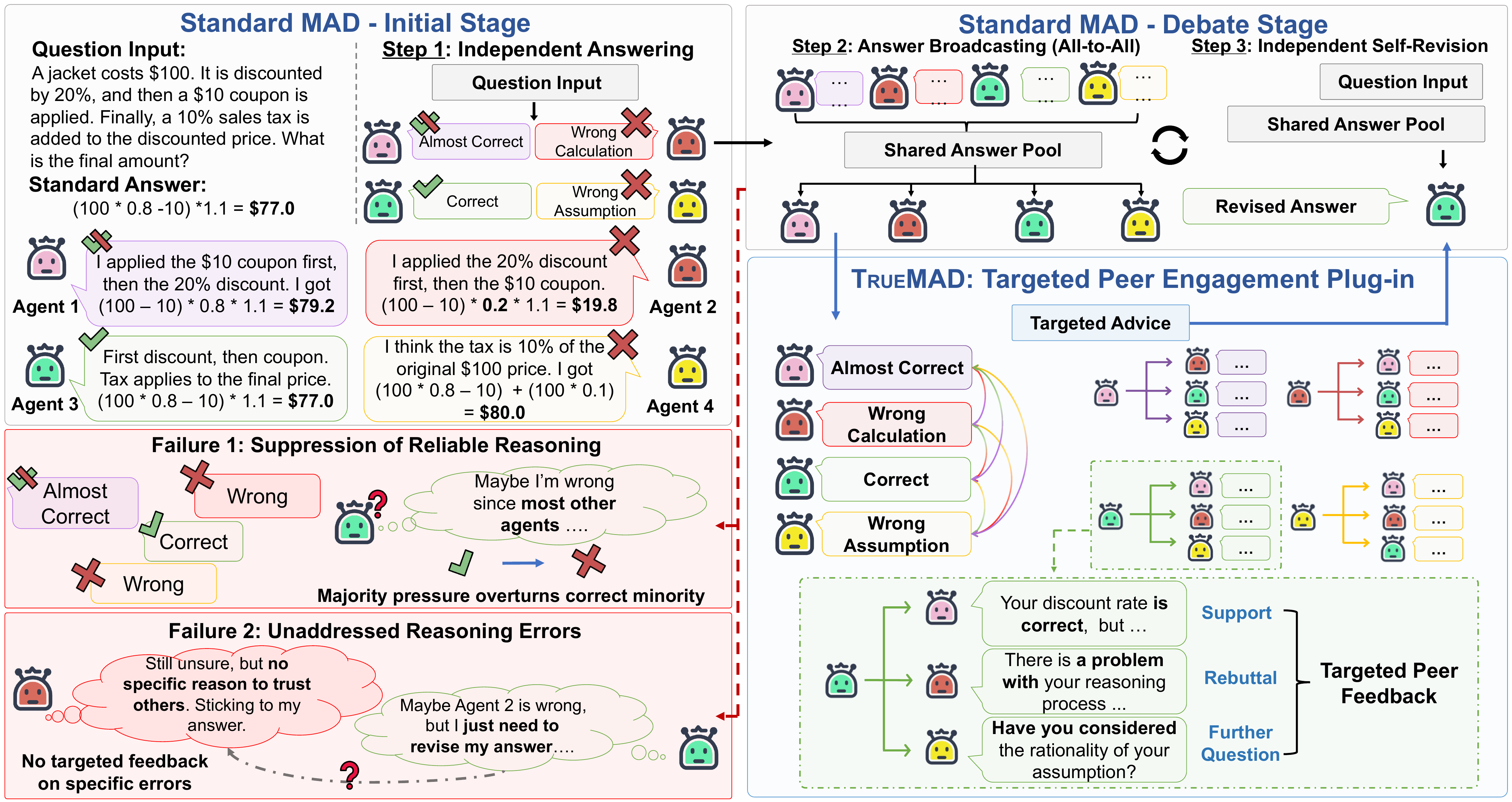}
  % \vspace{-0.3cm}
  \caption{Motivating example and \textsc{TrueMAD} overview. \textbf{Left:} In broadcast-only MAD, majority pressure can suppress a correct answer, while agents with distinct errors receive no targeted diagnosis. \textbf{Right:} \textsc{TrueMAD} adds one-to-one, free-form advice between broadcasting and self-revision, enabling support, rebuttal, refinement, or questions without altering the MAD pipeline.}  
  \label{figure-main-figure}
  \vspace{-0.4cm}
\end{figure}

However, whether MAD benefits from this advantage depends on how agents engage during debate. Standard MAD broadcasts previous-round responses to all agents and asks each to update'' or revise'' its answer~\citep{yin2023exchange}. At the outcome level, this setup is closer to answer sharing than targeted deliberation: agents first provide independently sampled answers, and reported gains may largely reflect repeated sampling and majority voting rather than peer reasoning~\citep{choi2025debateorvote,smit2024should,wang2024rethinking}. Other studies show that unconstrained peer influence can induce conformity and echo-chamber dynamics: agents may imitate apparent majorities, reinforce shared misconceptions, and suppress correct minority reasoning, increasing consensus without reliably improving performance~\citep{estornell2024multi,oh2025debate}. Thus, exchanging more answers does not ensure effective deliberation. As Figure~\ref{figure-main-figure} illustrates, this limitation can manifest in two complementary ways: reliable reasoning may be suppressed during revision, while agents that need correction may receive no targeted feedback on their errors.

First, \textbf{broadcast-style revision can suppress reasoning that is already
correct}. Prior work shows that shared misconceptions and majority influence
can create echo-chamber dynamics, reinforce erroneous beliefs, and even
overturn initially correct answers~\citep{estornell2024multi,oh2025debate,
choi2025debateorvote}. A shared answer pool presents competing responses
without explicitly distinguishing the reasoning that should be preserved from
the reasoning that should be reconsidered. A generic revision instruction also
gives no peer the responsibility to examine and support a valid derivation or
to rebut an erroneous majority on its behalf. As illustrated by \textit{Failure 1} in Figure~\ref{figure-main-figure}, one agent has already derived the
correct answer, whereas the other agents make different computational or
conceptual errors. Nevertheless, the number of conflicting responses may
become a stronger revision signal than the quality of the correct derivation,
causing the correct agent to abandon its answer. Once such reasoning
disappears from the debate trajectory, subsequent interaction and final
aggregation can amplify the resulting error rather than recover the lost
signal. Standard MAD may therefore increase consensus while weakening the
reasoning that interaction should preserve.

Second, \textbf{broadcast answer sharing can leave reasoning errors unaddressed even
when multiple peer responses are available}. Broader evaluations show that
adding agents, rounds, or alternative debate strategies does not guarantee
consistent improvements, while recent diagnostic studies caution that answer
changes and final consensus may reflect instability or social influence rather
than genuine correction~\citep{smit2024should,wang2024rethinking,
hao2026not,wang2026consistency,cui2026free}. One reason is that every agent
receives the same pool of answers, but no agent is explicitly asked to inspect
a specific peer response, locate its error, and explain how that error affects
the conclusion. As illustrated by \textit{Failure 2} in Figure~\ref{figure-main-figure}, the almost-correct answer,
the incorrect calculation, and the incorrect assumption require different
forms of feedback. Simply displaying the correct response does not explain to
each agent which step should be revised or why. Consequently, agents may retain
their original errors, make an ungrounded change, or repeatedly exchange
answers without resolving their disagreement. Together, the two failures
expose the same interaction gap: standard MAD makes peer answers visible but
lacks an explicit mechanism for transforming answer exposure into targeted
peer engagement. Agents may therefore fail both to defend reliable reasoning
and to provide the specific feedback needed to correct unreliable reasoning.

These limitations motivate a principle: before revising its answer,
each agent should have an explicit opportunity to examine and respond to the
reasoning of a specific peer. Based on this principle, we propose
\textsc{TrueMAD}, a backbone-agnostic test-time plug-in that introduces
one-to-one, free-form peer advice before self-revision. Each source agent
generates advice conditioned on a target agent's response, and the target then
revises using its own response, the shared peer-answer pool, and the advice
addressed to it. The advice may support a valid derivation, identify a specific
error, refine an incomplete argument, or raise a relevant question. Thus,
\textsc{TrueMAD} makes the source and target of interaction explicit without
forcing agents into adversarial roles. It requires no parameter updates and is
designed as a general plug-in for standard MAD and its improved variants. By
enabling agents to support, challenge, and refine specific peer reasoning,
\textsc{TrueMAD} seeks to turn answer sharing into effective interaction and
thereby promote collective intelligence through debate.

We empirically evaluate \textsc{TrueMAD} across multiple reasoning benchmarks
and show that it improves both standard MAD and an improved MAD variant when
used as a plug-in. Round-wise and difficulty-conditioned analyses further
examine when targeted interaction is most useful. To understand how it affects
the debate process, we observe the resulting dynamics at three progressively
broader granularities: peer utterances, individual-agent answer transitions,
and group-level consensus evolution. Additional call-controlled and
cross-model experiments provide further evidence for the effectiveness and
general applicability of targeted peer engagement.

To summarize, our contributions are threefold:
\begin{itemize}
    \item \textbf{A failure-driven motivation for targeted interaction.} We identify two limitations of broadcast-style MAD---the suppression
    of reliable reasoning and the lack of targeted correction---and motivate
    explicit peer engagement (Section~\ref{sec:truemad-motivation}).

    \item \textbf{A targeted-deliberation plug-in.} We propose \textsc{TrueMAD}, a backbone-agnostic test-time plug-in that introduces one-to-one, free-form peer advice before self-revision (Section~\ref{section-methods}).

    \item \textbf{Multi-granularity empirical evidence.} We validate \textsc{TrueMAD} across tasks and models, and analyze
    debate dynamics at the utterance, agent-transition, and group levels
    (Section~\ref{section-experiments}).
\end{itemize}

\section{Related Work}
\label{section-related-work}

\paragraph{Multi-Agent Debate and Protocol Variants.}
Multi-agent debate (MAD) is an inference-time collaboration paradigm in which
multiple LLM agents independently generate answers, exchange responses, and
revise them before final aggregation. Early work established this paradigm and
showed that interaction among agents can improve reasoning and factuality
without parameter updates~\citep{yin2023exchange,du2024improving,
liang2024encouraging}. Subsequent studies improve MAD through diverse
reasoning strategies and roles, confidence-aware consensus, grouped
communication, structured criticism, dynamic contexts, and internalized
debate~\citep{chan2024chateval,chen2024reconcile,liu2025breaking,
liu2024groupdebate,estornell2024multi,hua2026context,li2026debate,
yi2026latent}. These methods primarily improve agent diversity, communication
structure, context organization, or decision making. Yet most retain answer
exchange as the primary interaction unit without requiring claim-specific
feedback before self-revision. We focus on this interaction gap:
\textsc{TrueMAD} introduces one-to-one, free-form peer advice as a general
plug-in for MAD and its variants.

\paragraph{Re-examining Interaction in MAD.}
Recent studies question whether MAD gains consistently arise from substantive
deliberation rather than repeated sampling, majority aggregation, or
unconstrained peer influence. Empirical and theoretical analyses show that
voting can explain a substantial portion of reported gains, that performance
is sensitive to debate and decision protocols, and that shared misconceptions
or majority influence can produce echo chambers and overturn initially correct
answers~\citep{smit2024should,wang2024rethinking,choi2025debateorvote,
estornell2024multi,oh2025debate}. Other work cautions that answer changes may
reflect spontaneous instability rather than genuine peer-induced correction,
while answer-level consensus can conceal disagreement in the underlying
reasoning~\citep{hao2026not,wang2026consistency}. Consensus-free approaches
accordingly explore decision rules that do not rely solely on final
agreement~\citep{cui2026free}. Together, these findings align with the two
limitations highlighted in our Introduction: reliable reasoning may be
suppressed during revision, while erroneous reasoning may remain uncorrected
when answer exposure does not provide claim-specific feedback. Complementary
to work that diagnoses outcomes or redesigns decision rules, we connect both
failures to the interaction design, introduce targeted peer advice, and
examine its effects at the utterance, agent-transition, and group levels.

\section{Methods: TrueMAD as a Plug-in Interaction Protocol}
\label{section-methods}

\subsection{Motivation}
\label{sec:truemad-motivation}

The limitation of broadcast-style MAD is not that agents lack access
to peer answers, but that answer exposure does not define how agents should
engage with one another's reasoning. As discussed in
Sections~\ref{section-intro} and~\ref{section-related-work}, this interaction
gap can lead to two complementary failures. Reliable reasoning may be
suppressed when an agent revises under the influence of conflicting or
majority responses, while erroneous reasoning may remain uncorrected because
no peer is responsible for locating and explaining the error. Both
failures arise between answer broadcasting and self-revision: the protocol
provides a shared answer pool, but does not turn peer reasoning into feedback
directed at a particular recipient. Our motivation is therefore not to
increase communication or disagreement, but to make peer engagement explicit
before each agent decides whether and how to revise.

Several intuitive ways of strengthening standard MAD do not directly
resolve this interaction gap. Increasing the number of agents or debate
rounds expands answer exposure, but does not ensure that any agent examines
a specific peer claim. Strengthening a generic revision prompt may elicit
more elaborate responses while leaving the source and target of feedback
implicit. Imposing fixed adversarial roles guarantees opposition rather than
evidence-sensitive engagement, and may unnecessarily challenge valid
reasoning. These limitations motivate three design goals.

\paragraph{Goal 1: Targeted and Actionable Engagement.}
Each feedback message should have an explicit source and target, conditioned on
the target's current response. This requirement gives the source agent a
concrete object to inspect: it can identify a claim, reasoning step,
assumption, or calculation and explain how it should affect the target's next
revision. Because the protocol cannot know which agent holds reliable
reasoning, we instantiate this interaction over ordered agent pairs rather
than relying on an external judge or routing decision. Target specificity
turns a shared pool of answers into recipient-oriented feedback rather than
another undirected group summary.

\paragraph{Goal 2: Flexible Support and Correction.}
Effective deliberation should not equate interaction with opposition. A peer
response may be correct, partially correct, or wrong, so feedback should
remain free-form: it may support a valid derivation, rebut a specific error,
refine an incomplete argument, or raise a relevant question. This addresses
both failures. Agents can receive reasons to retain well-supported reasoning
as well as guidance for reconsidering unreliable reasoning, allowing
interaction to promote collective intelligence without imposing artificial
disagreement.

\paragraph{Goal 3: Recipient-controlled and Protocol-compatible Revision.}
Peer feedback should inform rather than replace self-revision. The recipient
should revise using its own response, the broadcast peer answers, and the
advice addressed to it, while remaining free to accept, reject, or partially
use that advice. The intervention should also require no parameter updates,
fixed adversarial roles, or external judge at inference, and should be
applicable to standard MAD and its improved variants without replacing their
underlying reasoning procedures. We therefore realize these goals as a
test-time plug-in that inserts a one-to-one advice stage immediately before
self-revision.

\subsection{MAD Formulation}
\label{sec:mad-formulation}

Following standard MAD protocols~\citep{du2024improving,liang2024encouraging},
let $\mathcal{B}$ denote a base protocol with $N$ agents and $T$ response
rounds. Given a question $q$, agent $i$ produces its initial response as
\begin{equation}
x_i^1=S_{\mathcal{B},i}(q),
\qquad i=1,\ldots,N,
\label{eq:mad-initialization}
\end{equation}
where $S_{\mathcal{B},i}$ is the protocol-specific initializer. Let
$\mathcal{X}^t=(x_1^t,\ldots,x_N^t)$ denote the response collection at round
$t$, and let $\mathcal{X}_{-i}^t=(x_j^t)_{j\neq i}$ contain the peer responses
available to agent $i$. Each debate-stage update is
\begin{equation}
x_i^{t+1}
=
U_{\mathcal{B},i}
\left(q,x_i^t,\mathcal{X}_{-i}^t\right),
\qquad t=1,\ldots,T-1,
\label{eq:mad-update}
\end{equation}
and the final group prediction is
\begin{equation}
\hat{y}=A_{\mathcal{B}}\!\left(\mathcal{X}^T\right),
\label{eq:mad-aggregation}
\end{equation}
where $U_{\mathcal{B},i}$ and $A_{\mathcal{B}}$ are the base protocol's
updater and aggregation operator, respectively. The agent-indexed operators
cover both homogeneous MAD and variants with agent-specific reasoning
procedures. We use
$\operatorname{ans}(x_i^t)$ to denote the final answer extracted from a
response.

\subsection{\textsc{TrueMAD}: Targeted Peer Engagement}
\label{sec:truemad}

As shown in the right half of Figure~\ref{figure-main-figure},
\textsc{TrueMAD} preserves standard MAD pipeline while inserting a
targeted advice stage between answer broadcasting and self-revision. At round
$t$, agents first share response collection $\mathcal{X}^t$ as in base
protocol. \textsc{TrueMAD} then generates advice from source agents
to targets, after which each target revises using
the broadcast responses and the advice addressed to it. The plug-in therefore
supplements, rather than replaces, the original answer-sharing path.

\paragraph{Targeted and Flexible Advice.}
Let
\[
\mathcal{E}_N
=
\{(i,j):i,j\in\{1,\ldots,N\},\ i\neq j\}
\]
denote the ordered source--target pairs. Because the protocol does not know in
advance which response contains reliable reasoning, advice is generated over
all ordered pairs rather than through an external router or judge. For each
$(i,j)\in\mathcal{E}_N$, source agent $i$ advises target agent $j$ as
\begin{equation}
c_{i\rightarrow j}^{t}
=
G\!\left(q,x_i^t,x_j^t\right),
\qquad t=1,\ldots,T-1.
\label{eq:truemad-advice-generation}
\end{equation}
Conditioning on both responses gives the source a reasoning path to
examine. It can address a claim, step, assumption, or calculation in the
target response instead of producing a generic summary of the group. The
advice is free-form: as illustrated in
Figure~\ref{figure-main-figure}, it may support a sound derivation, rebut a
specific error, refine an incomplete argument, or raise a question.
Thus, interaction is target-specific without being forced into a fixed
adversarial role. This design allows reliable reasoning to receive explicit
support while providing erroneous reasoning with feedback suited to its
weakness.

\paragraph{Recipient-oriented Revision.}
After pairwise generation, advice is grouped by target. Agent $j$ receives
\begin{equation}
\mathcal{C}_j^t
=
\left(\left(i,c_{i\rightarrow j}^{t}\right)\right)_{i\neq j},
\label{eq:truemad-received-advice}
\end{equation}
where each message retains its source identity. Thus, agents share the broadcast responses but receive target-specific advice conditioned on their current responses. Each target then updates as
\begin{equation}
x_j^{t+1}
=
U_{\mathcal{B},j}^{+}
\left(q,x_j^t,\mathcal{X}_{-j}^t,\mathcal{C}_j^t\right),
\label{eq:truemad-update}
\end{equation}
where $U_{\mathcal{B},j}^{+}$ denotes the base updater
$U_{\mathcal{B},j}$ augmented with $\mathcal{C}_j^t$ in its context. The
recipient retains its response and broadcast peer responses, free to accept,
reject, reconcile, or partially use the advice. Advice guides self-revision
rather than serving as an intermediate vote or hard revision command.

This procedure repeats before each debate-stage revision. It requires no
parameter updates, fixed adversarial roles, or external judge, and preserves
the base protocol's initialization, reasoning, and final aggregation. For
standard MAD, advice is added to its revision context; for variants with
agent-specific or multi-step updaters, it is supplied to the existing
procedure. \textsc{TrueMAD} therefore implements the three design goals
through targeted generation, flexible feedback, and recipient-controlled
revision, while remaining applicable to MAD and improved variants.

\section{Experiments}
\label{section-experiments}

In this section, we evaluate \textsc{TrueMAD} by addressing the following research questions:

\textbf{RQ1.} To what extent does \textsc{TrueMAD} improve the performance of existing MAD protocols?

\textbf{RQ2.} Under which debate conditions does \textsc{TrueMAD} yield the greatest performance gains?

\textbf{RQ3.} How does \textsc{TrueMAD} influence debate dynamics?

\textbf{RQ4.} Are \textsc{TrueMAD}'s gains attributable to additional inference or to peer engagement?

\textbf{RQ5.} How robust are the performance gains of \textsc{TrueMAD} across different LLM backbones?

\subsection{Experimental Setup}
\label{section-experiments-setup}

\textbf{Backbone and Datasets.} The main experiments use \texttt{Qwen3-30B-A3B}~\citep{qwen3technicalreport} as the backbone model. We evaluate
multi-agent debate on three complementary benchmarks: AIME~\citep{dekoninck2026matharena}, GPQA-Diamond~\citep{rein2023gpqa}, and TruthfulQA~\citep{lin2022truthfulqa}. AIME combines the 2024, 2025, and 2026 test sets for the
primary comparison, while GPQA-Diamond and TruthfulQA evaluate scientific
question answering and truthfulness, respectively. Following prior multi-agent debate studies that explicitly account for the computational cost of multi-agent evaluation~\citep{wang2024rethinking,smit2024should, chen2024reconcile}, we randomly sample 200 questions from TruthfulQA using a fixed random seed of 0.

\textbf{Baselines.} We compare five protocols: MAD, promptMAD, DMAD, \textsc{TrueMAD}, and \textsc{TrueDMAD}. MAD is the vanilla broadcast-only multi-agent debate baseline. promptMAD uses the same generic debate instruction as \textsc{TrueMAD} but without introducing directed advice. DMAD is a diversity-oriented MAD variant that changes the agent-level reasoning conditions~\citep{liu2025breaking}. \textsc{TrueMAD} adds the one-to-one free-form advice stage described in Section~\ref{sec:truemad} to vanilla MAD, whereas \textsc{TrueDMAD} adds the same advice stage to DMAD.

\textbf{Debate Protocol.} All main experiments use four agents and five response rounds. Round 1 contains independent initial responses, and rounds 2--5 contain successive debate-stage updates. For each matched comparison, we keep the backbone, questions, initial-stage sampling prompts, agent count, number of rounds, answer extraction procedure, and final aggregation rule fixed. 

\textbf{Metric.}
We evaluate fine-grained agent-level predictions and the final aggregated group
answer. For fine-grained evaluation, we report agent-level accuracy
(\textbf{Agent ACC}). Let
$\mathcal{D}={(q_n,y_n)}_{n=1}^{|\mathcal{D}|}$ be a dataset, where
$q_n$ and $y_n$ denote question $n$ and its reference answer, respectively.
Let $x_{n,i}^{t}$ denote the response of agent $i$ to question $q_n$ at round
$t$. Agent-level accuracy is
\begin{equation}
\mathrm{ACC}^{t}
=
\frac{1}{|\mathcal{D}|N}
\sum_{n=1}^{|\mathcal{D}|}
\sum_{i=1}^{N}
\mathbf{1}\!\left[
\operatorname{ans}(x_{n,i}^{t})=y_n
\right],
\qquad
t=1,\ldots,T.
\label{eq:agent-level-accuracy}
\end{equation}
Here, $N$ is the number of agents and $T$ is the number of response rounds.

\subsection{Experimental Results (RQ1 \& RQ2)}
\label{section-experiments-analysis}

\begin{table*}[t]
    \centering
    \caption{Main results on AIME, GPQA-Diamond, and TruthfulQA with \texttt{Qwen3-30B-A3B}. We report Agent ACC (\%) and Vote ACC (\%)  as mean $\pm$ sample standard deviation over three seeds. Bold marks the best method in each column; $\Delta$ rows show the gains from adding \textsc{TrueMAD}. }
    \label{tab:qwen3-30b-main-results}
    \setlength{\tabcolsep}{2pt}
    \begin{tabular*}{\textwidth}{@{\extracolsep{\fill}}lcccccc@{}}
    \hline
    \multirow{2}{*}{\textbf{Method}}
    &
    \multicolumn{2}{c}{AIME}
    & \multicolumn{2}{c}{GPQA-Diamond}
    & \multicolumn{2}{c}{TruthfulQA} \\
    \cline{2-3}
    \cline{4-5}
    \cline{6-7}
    & Agent ACC & Vote ACC & Agent ACC & Vote ACC & Agent ACC & Vote ACC \\
    \hline
    Single LLM
    & \multicolumn{2}{c}{$\acc{23.06}{1.27}$} 
    & \multicolumn{2}{c}{$\acc{52.82}{0.07}$}
    & \multicolumn{2}{c}{$\acc{77.17}{0.19}$} \\
    \hline
    MAD
    & $\acc{29.81}{2.49}$ & $\acc{29.63}{2.57}$
    & $\acc{54.12}{1.29}$ & $\acc{52.86}{1.27}$
    & $\acc{77.88}{0.45}$ & $\acc{76.50}{1.32}$ \\
    promptMAD
    & $\acc{30.37}{2.39}$ & $\acc{30.00}{2.94}$
    & $\acc{55.26}{1.52}$ & $\acc{54.71}{1.77}$
    & $\acc{77.96}{0.29}$ & $\acc{76.83}{0.76}$ \\
    DMAD
    & $\acc{30.28}{1.00}$ & $\acc{29.63}{1.28}$
    & $\acc{56.14}{1.53}$ & $\acc{56.06}{1.82}$
    & $\acc{77.79}{1.25}$ & $\acc{77.33}{0.76}$ \\
    \textbf{TrueMAD}
    & $\accbold{35.37}{2.36}$ & $\accbold{34.81}{1.70}$
    & $\acc{58.00}{0.60}$ & $\acc{56.73}{0.29}$
    & $\acc{79.29}{0.62}$ & $\acc{78.00}{0.87}$ \\
    \textbf{TrueDMAD}
    & $\acc{34.63}{1.12}$ & $\acc{34.44}{1.11}$
    & $\accbold{62.25}{1.79}$ & $\accbold{60.27}{3.21}$
    & $\accbold{81.50}{0.38}$ & $\accbold{80.50}{1.32}$ \\
    \hline
    \textbf{$\Delta$ MAD}
    & $\mathbf{+5.56}$ & $\mathbf{+5.18}$
    & $\mathbf{+3.87}$ & $\mathbf{+3.87}$
    & $\mathbf{+1.42}$ & $\mathbf{+1.50}$ \\
    \textbf{$\Delta$ DMAD}
    & $\mathbf{+4.35}$ & $\mathbf{+4.81}$
    & $\mathbf{+6.10}$ & $\mathbf{+4.21}$
    & $\mathbf{+3.71}$ & $\mathbf{+3.17}$ \\
    \hline
    \end{tabular*}
\end{table*}

\paragraph{Comparative Results.} To answer RQ1, we evaluate whether targeted peer advice improves individual-agent predictions and the final aggregated group answer. Averaged across benchmarks, \textsc{TrueMAD} improves over MAD by $3.62$ percentage points in Agent ACC and $3.52$ points in Vote ACC; \textsc{TrueDMAD} improves over DMAD by $4.72$ and $4.06$ points, respectively. These improvements in Agent ACC and Vote ACC demonstrate that \textsc{TrueMAD} is effective not only at improving individual agents' responses but also at strengthening the aggregated output. The gains over both MAD and DMAD show that the advice stage can be integrated with homogeneous and strategy-diverse debate frameworks, rather than depending on an underlying reasoning procedure. promptMAD remains close to MAD on both metrics. This contrast suggests that the gains arise from explicit source--target peer engagement rather than from more elaborate revision wording alone.

\begin{figure}[t]
  \centering
  \includegraphics[width=\linewidth]{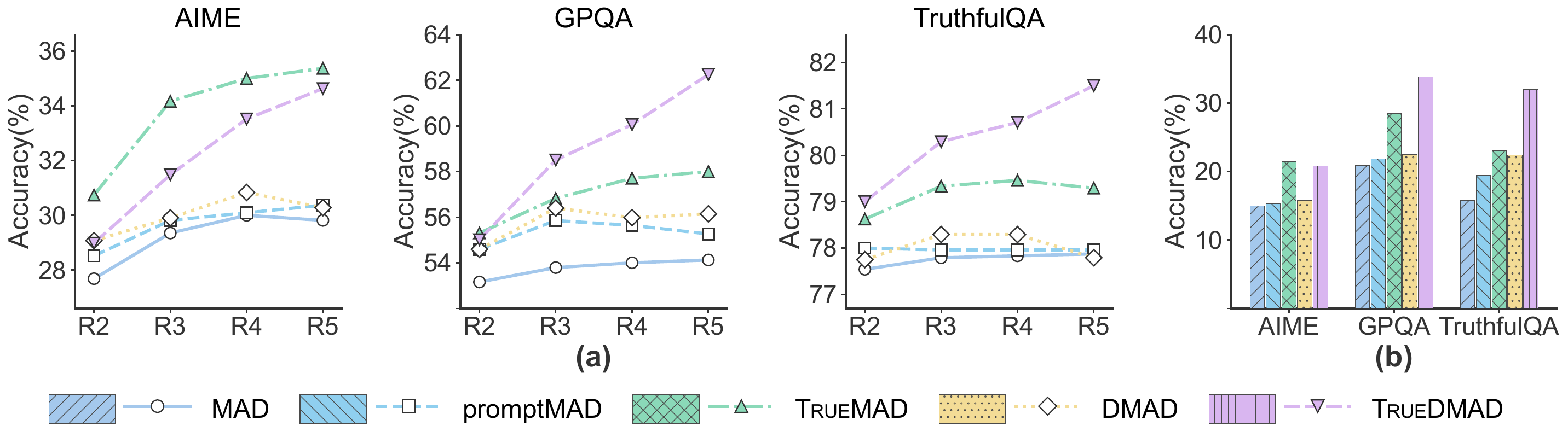}
  \vspace{-0.3cm}
  \caption{Agent-level accuracy (\%) across debate rounds and difficult initial conditions with
    \texttt{Qwen3-30B-A3B}. \textbf{(a)} Round-wise accuracy from R2 to R5 on AIME,
    GPQA-Diamond, and TruthfulQA. \textbf{(b)} R5 accuracy on questions for which at
    least half of the four agents are incorrect in R1.}
  \label{fig:qwen3-30b-acc-panel}
  \vspace{-0.4cm}
\end{figure}

\paragraph{Further Gains with Debate Rounds.} Figure~\ref{fig:qwen3-30b-acc-panel}(a) shows different protocol trajectories. MAD typically gains in the early rounds and then approaches a plateau, whereas \textsc{TrueMAD} continues to improve through R5 on AIME and GPQA-Diamond. From R3 to R5, \textsc{TrueMAD} gains $1.20$ points on AIME and $1.18$ on GPQA-Diamond, compared with $0.46$ and $0.33$ for MAD. The difference is even more pronounced for \textsc{TrueDMAD}: it gains $3.15$ and $3.75$ points from R3 to R5, while DMAD is nearly flat. These trajectories indicate that \textsc{TrueMAD} does not merely provide an early accuracy boost or collapse into premature consensus. A plausible explanation is that peer-specific advice keeps later rounds focused on evaluating individual claims, allowing agents to correct or defend an answer instead of repeatedly following the current majority. The smaller late-round change for \textsc{TrueMAD} is consistent with the higher initial accuracy on TruthfulQA, where many questions are already answered correctly by most agents at R1 and leave less room for further improvement.

\paragraph{Larger Gains in Difficult Scenarios.}
To answer RQ2, we examine questions for which at least half of the agents
answer incorrectly in R1, as shown in
Figure~\ref{fig:qwen3-30b-acc-panel}(b). In this regime, the initial group
lacks a reliable correct majority. Averaged equally across the three
benchmarks, \textsc{TrueMAD} exceeds MAD by $7.13$ points in R5 accuracy,
while \textsc{TrueDMAD} exceeds DMAD by $8.63$ points. The pattern is consistent with the intended role of the
interaction: when the broadcast majority is unreliable, a directed response
can help agents inspect the evidence behind individual answers and preserve a
correct minority signal that would otherwise be overlooked. This does not
mean that advice can reliably recover an answer when all agents are wrong;
rather, it is most useful when the initial debate contains partial but
unevenly distributed evidence.

\subsection{Debate Dynamics Analysis (RQ3)}
\label{section-experiments-debate-dynamics}

To answer RQ3, we observe the debate process at three granularities: the
\emph{utterance level}, which examines what agents communicate; the
\emph{agent-transition level}, which tracks how answers change;
and the \emph{group level}, which measures how correct and erroneous
consensus evolves. We use the matched comparisons \textsc{TrueMAD} versus MAD
and \textsc{TrueDMAD} versus DMAD throughout. These quantities are
defined for a single question; we suppress the dataset index $n$ for
readability.

\paragraph{Level 1: Utterance-level Peer Deliberation.}
Our first objective is to determine whether interaction supplies substantive
peer reasoning before asking whether the recipient changes its answer. We
consider a message useful when it addresses a concrete peer claim, supports its
stance through reasoning rather than social agreement, and contributes
information that can guide the next revision. Accordingly, for each
debate-stage reply, an LLM judge returns
\begin{equation}
\boldsymbol{\phi}(x_i^t)
=
\bigl(
\phi_{\mathrm{RP}}(x_i^t),
\phi_{\mathrm{EI}}(x_i^t),
\phi_{\mathrm{NC}}(x_i^t)
\bigr),
\label{eq:peer-quality}
\end{equation}
where Rebuttal Precision (RP) measures engagement with a specific argument and
its error, Evidence Independence (EI) measures reasoning independent of
consensus or imitation, and Novel Contribution (NC) measures added reasoning
steps, alternatives, conditions, or checks. We use
\texttt{DeepSeek-V4-Flash}~\citep{xu2026deepseek} as the judge; the full rubric
is provided in Appendix~\ref{app:level1-judge-prompt}.

\begin{figure}[t]
  \centering
  \includegraphics[width=\linewidth]{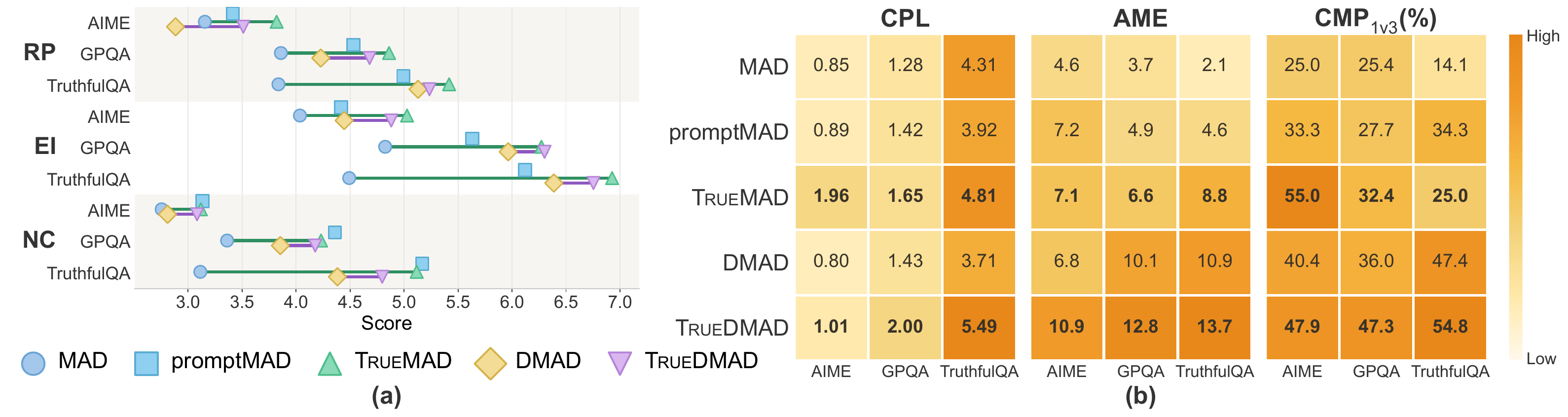}
  % \vspace{-0.3cm}
  \caption{Debate dynamics analysis of debate quality with \texttt{Qwen3-30B-A3B}. \textbf{(a)} LLM-judged utterance quality (1--10); lines connect MAD with \textsc{TrueMAD} and DMAD with \textsc{TrueDMAD}. \textbf{(b)} Agent- and group-level diagnostics. Higher is better for all metrics.}
  \label{fig:process-level-metrics}
  \vspace{-0.4cm}
\end{figure}

\noindent\textit{Results.}
Figure~\ref{fig:process-level-metrics}(a) shows uniform improvement across
comparisons: both \textsc{TrueMAD}-augmented methods
outperform their baselines on RP, EI, and NC for every benchmark.
The gain is not confined to producing more novel content; targeted
advice makes responses more specific to peer arguments and
more independently reasoned. Compared with promptMAD, \textsc{TrueMAD}
achieves higher RP and EI on all three benchmarks, whereas promptMAD is
slightly higher on NC. This contrast indicates that generic prompting mainly
encourages additional ideas, while target-specific advice
anchors those ideas to the peer claim and its supporting evidence.

\paragraph{Level 2: Agent-transition Correction Preference.}
Our second objective is to determine whether interaction helps wrong agents recover without
unnecessarily overturning correct agents. Raw flip counts are insufficient
because methods with different numbers of wrong and correct agents have
different opportunities for each transition. We therefore condition on
previous correctness state. For a question with reference answer $y$, let
$s_i^t=\mathrm{R}$ if $\operatorname{ans}(x_i^t)=y$, and let
$s_i^t=\mathrm{W}$ otherwise. We define
\begin{equation}
\begin{aligned}
p_{\mathrm{W}\rightarrow\mathrm{R}}
&=\Pr(s_i^t=\mathrm{R}\mid s_i^{t-1}=\mathrm{W}), &
p_{\mathrm{R}\rightarrow\mathrm{W}}
&=\Pr(s_i^t=\mathrm{W}\mid s_i^{t-1}=\mathrm{R}), &
\mathrm{CPL}
&=\frac{p_{\mathrm{W}\rightarrow\mathrm{R}}}
{p_{\mathrm{R}\rightarrow\mathrm{W}}}. 
\end{aligned}
\label{eq:correction-preference-lift}
\end{equation}
We pool eligible question--agent transitions across debate-stage transitions.
The resulting Correction Preference Lift (CPL) compares the propensity to
correct a wrong answer against the propensity to overturn a correct one; a
larger value indicates a more correction-oriented revision process.

\noindent\textit{Results.}
Figure~\ref{fig:process-level-metrics}(b) shows a consistent increase in CPL
across all six matched comparisons, indicating that targeted advice improves
the selectivity of revision rather than merely inducing more answer changes.
The effect is most revealing on AIME: MAD and DMAD both have CPL below $1$,
meaning that a correct agent is conditionally more likely to be overturned
than a wrong agent is to recover. Adding targeted advice raises these values
from $0.85$ to $1.96$ and from $0.80$ to $1.01$, respectively, reversing or
nearly eliminating this unfavorable imbalance. On GPQA-Diamond and
TruthfulQA, where the base protocols already favor correction, targeted advice
further strengthens that preference. The same directional effect under both
MAD and DMAD shows that the improvement is not tied to a particular
initialization strategy: targeted peer engagement systematically redirects
revision toward error correction and away from the loss of correct answers.

\paragraph{Level 3: Group-level Consensus Robustness.}
We next determine whether the improved transition balance yields more reliable collective dynamics. At the group level, disrupting consensus is not always beneficial: an erroneous majority should be broken to enable correction, whereas a correct majority should be preserved to avoid discarding existing correct evidence. We ask whether a protocol selectively disrupts consensus---encouraging agents to leave erroneous majorities while retaining correct ones---rather than simply destabilizing the group. Let $m^{t-1}$ be the unique modal answer when at least three of four agents support it, and let $M^{+}$ and $M^{-}$ denote correct- and erroneous-majority states. For agents supporting the previous majority, we define
\begin{equation}
\begin{aligned}
E^t(M)
&=\Pr\!\left(
\operatorname{ans}(x_i^t)\neq m^{t-1}
\mid
\operatorname{ans}(x_i^{t-1})=m^{t-1},M
\right),\\
\mathrm{AME}
&=\frac{1}{T-1}\sum_{t=2}^{T}
\left(E^t(M^{-})-E^t(M^{+})\right).
\end{aligned}
\label{eq:asymmetric-majority-escape}
\end{equation}
Asymmetric Majority Escape (AME) separates selective resistance to erroneous
consensus from general instability. Because leaving an erroneous majority does
not necessarily yield the correct answer, we interpret AME jointly with CPL
and with Correct-Minority Propagation,
\begin{equation}
\mathrm{CMP}_{1v3}
=
\Pr(k^t>1\mid k^{t-1}=1),
\label{eq:correct-minority-propagation}
\end{equation}
where $k^t=\sum_{i=1}^{N}\mathbf{1}[s_i^t=\mathrm{R}]$ is the number of
correct agents at round $t$. CMP$_{1v3}$ measures
whether evidence initially held by the sole correct agent spreads to a peer.

\noindent\textit{Results.}
Figure~\ref{fig:process-level-metrics}(b) shows that AME and CMP$_{1v3}$ improve
together in every matched comparison. Their joint improvement addresses two
complementary group-level failures: the higher AME reflects a more favorable
balance between challenging erroneous consensus and preserving correct
consensus, while the higher CMP$_{1v3}$ shows that a correct minority is less
likely to be suppressed and more likely to recruit additional agents.
\textsc{TrueMAD} therefore does not simply make agents less conformist or the
group more volatile. Instead, it makes consensus formation more
evidence-responsive: harmful consensus is more readily reconsidered, whereas
correct evidence is more likely to survive and spread. Together with the
higher CPL, these results connect correction-oriented individual revisions to
more reliable collective outcomes.

Across the three granularities, \textsc{TrueMAD} improves the reasoning
available to recipients, shifts individual revisions toward correction, and
produces more favorable group-level consensus dynamics. 

\subsection{Inference Budget and Peer Engagement (RQ4)}
\label{section-experiments-call-matched}

\begin{wraptable}{r}{0.5\textwidth}
    \centering
    \caption{Auxiliary-call-matched comparison with \texttt{Qwen3-30B-A3B}. Bold marks the best method.}
    \label{tab:call-matched-controls}
    \scriptsize
    \setlength{\tabcolsep}{2pt}
    \begin{tabular*}{0.5\textwidth}{@{\extracolsep{\fill}}lccc@{}}
    \hline
    \textbf{Method} & \textbf{AIME} & \textbf{GPQA-Diamond} & \textbf{TruthfulQA} \\
    \hline
    MAD
    & $\acc{29.81}{2.49}$
    & $\acc{54.12}{1.29}$
    & $\acc{77.88}{0.45}$ \\
    \hline
    ComputeMAD
    & $\acc{31.48}{1.05}$
    & $\acc{56.40}{2.63}$
    & $\acc{77.08}{1.18}$ \\
    Self-AdviceMAD
    & $\acc{33.33}{1.00}$
    & $\acc{54.55}{2.70}$
    & $\acc{75.21}{1.34}$ \\
    \textbf{TrueMAD}
    & $\accbold{35.37}{2.36}$
    & $\accbold{58.00}{0.60}$
    & $\accbold{79.29}{0.62}$ \\
    \hline
    \end{tabular*}
\end{wraptable}

To answer RQ4, we examine whether the gains persist when the same
auxiliary-call budget is allocated to feedback without directed peer
engagement. \textit{ComputeMAD} generates non-targeted notes from the previous-round
group responses and randomly assigns three notes to each agent, whereas
\textit{Self-AdviceMAD} generates three independent critiques of each agent's own
response. Both variants use the same number of auxiliary calls and the same revision
input structure as \textsc{TrueMAD}. As shown in Table~\ref{tab:call-matched-controls},
\textsc{TrueMAD} achieves the highest mean R5 accuracy on every benchmark,
averaging $2.57$ points above \textit{ComputeMAD} and $3.19$ points above
\textit{Self-AdviceMAD}. The two controls exhibit task-dependent effects and both fall
below MAD on TruthfulQA, whereas \textsc{TrueMAD} improves over MAD across all
three tasks. Under the same
auxiliary-call count, this pattern suggests that additional inference alone is
insufficient to reproduce the gains and that the organization of feedback
into directed peer engagement is consequential. Because the variants use
independently generated R1 responses, we interpret this comparison as
supporting rather than definitive causal evidence.

\subsection{Cross-Model Stability (RQ5)}
\label{section-experiments-cross-model}

\begin{wraptable}{r}{0.45\textwidth}
    \centering
    \caption{Cross-model results on AIME. Bold marks the best method for each backbone.}
    \label{tab:gpt41mini-qwen332b-aime-results}
    \scriptsize
    \begin{tabular*}{0.45\textwidth}{@{\extracolsep{\fill}}lcc@{}}
    \hline
    \textbf{Method} & \textbf{\texttt{GPT-4.1-mini}} & \textbf{\texttt{Qwen3-32B}} \\
    \hline
    Single LLM
    & $\acc{51.57}{0.85}$ & $\acc{18.43}{0.42}$ \\
    \hline
    MAD
    & $\acc{62.87}{0.70}$ & $\acc{23.98}{1.60}$ \\
    promptMAD
    & $\acc{63.06}{1.21}$ & $\acc{24.91}{1.67}$ \\
    DMAD
    & $\acc{63.52}{0.42}$ & $\acc{24.72}{1.82}$ \\
    \textbf{TrueMAD}
    & $\acc{67.31}{1.79}$ & $\acc{27.69}{1.31}$ \\
    \textbf{TrueDMAD}
    & $\accbold{68.61}{2.20}$ & $\accbold{28.33}{2.27}$ \\
    \hline
    \textbf{$\Delta$ MAD}
    & $\mathbf{+4.44}$ & $\mathbf{+3.70}$ \\
    \textbf{$\Delta$ DMAD}
    & $\mathbf{+5.09}$ & $\mathbf{+3.61}$ \\
    \hline
    \end{tabular*}
\end{wraptable}

To answer RQ5, we evaluate the same debate protocol on
\texttt{Qwen3-32B}~\citep{qwen3technicalreport}, as a representative dense model, and
\texttt{GPT-4.1-mini}~\citep{openai2025gpt41mini}, as a representative closed-source
model, using AIME. Table~\ref{tab:gpt41mini-qwen332b-aime-results} shows that
\textsc{TrueMAD} achieves higher mean accuracy than MAD, while
\textsc{TrueDMAD} achieves higher mean accuracy than DMAD, on both backbones.
These results provide cross-model evidence on AIME that targeted peer advice
remains effective as a plug-in interaction operator across different reasoning
models.

\section{Conclusion}
\label{section-conclusion}

This work identifies an interaction gap in broadcast-style MAD: visible peer
answers do not ensure engagement, allowing reliable reasoning to be suppressed
and specific errors to remain unaddressed. We propose \textsc{TrueMAD}, a
test-time plug-in that introduces one-to-one, free-form advice before
self-revision. Across benchmarks, \textsc{TrueMAD} improves MAD and DMAD;
analyses at the utterance, agent-transition, and group levels show more
targeted deliberation and correction-oriented dynamics. These results suggest
that effective multi-agent reasoning must move beyond answer sharing toward
targeted peer engagement. Future work will explore selective advice to reduce cost.

\bibliography{iclr2027_conference}
\bibliographystyle{iclr2027_conference}

\appendix
\section{Experiment Setup Details}
\label{section-appendix-experiment-setup-details}

\subsection{Datasets}
\label{section-appendix-experiment-setup-details-datasets}
We evaluate the methods on three benchmarks widely used in prior studies~\citep{smit2024should, estornell2024multi, liu2025breaking, hua2026context, cui2026free, liu2026prepare}. We then explain each dataset separately.

\begin{itemize}
    \item \textbf{AIME}~\citep{dekoninck2026matharena}, is a benchmark based on challenging high-school mathematics competition problems with integer-valued answers, designed to evaluate advanced mathematical reasoning, creative problem-solving, and multi-step derivation. We used data from the years 2024, 2025, and 2026, for a total of 90 open-ended questions.
    \item \textbf{GPQA-Diamond}~\citep{rein2023gpqa}, is the highest-quality subset of GPQA, comprising 198 expert-validated graduate-level multiple-choice questions in biology, physics, and chemistry that assess deep domain knowledge and rigorous scientific reasoning.
    \item \textbf{TruthfulQA}~\citep{lin2022truthfulqa}, is a benchmark of questions designed to elicit common misconceptions and imitative falsehoods, evaluating whether models generate answers that are both truthful and informative across diverse domains. As mentioned in Section~\ref{section-experiments-setup}, we drew one fixed sample of 200 questions using random seed 0 to limit evaluation cost.
\end{itemize}

\subsection{Baselines}
\label{section-appendix-experiment-setup-details-baselines}

We compare five multi-agent debate protocols. For the published baselines,
we follow their publicly available implementations and integrate all methods
into a unified evaluation pipeline. Within each matched comparison, we keep
the backbone model, questions, number of agents,
number of debate rounds, answer extraction procedure, and final aggregation
rule fixed. The compared methods are as follows:

\begin{itemize}
    \item \textbf{MAD}~\citep{yin2023exchange,du2024improving,
    liang2024encouraging}.
    The standard multi-agent debate baseline first asks multiple homogeneous
    agents to solve a question independently. In each subsequent round, every
    agent observes the previous-round responses of all other agents and
    revises its own solution accordingly. Since the same pool of peer
    responses is broadcast to all agents, MAD does not explicitly specify
    which peer should evaluate or respond to a particular answer.

    \item \textbf{promptMAD}.
    A prompt-level control that preserves the broadcast communication
    topology and update schedule of MAD. It augments the revision prompt with
    a generic instruction encouraging agents to critically evaluate peer
    reasoning, but does not generate or route explicit peer-specific advice.
    This control helps determine whether any improvement can be attributed
    merely to stronger debate instructions rather than to the directed
    interaction mechanism introduced by \textsc{TrueMAD}.

    \item \textbf{DMAD}~\citep{liu2025breaking}.
    Diverse Multi-Agent Debate addresses homogeneous reasoning by assigning
    different problem-solving strategies to different agents. The agents
    examine solutions produced under these distinct reasoning conditions and
    iteratively revise their responses through debate. Compared with vanilla
    MAD, DMAD primarily modifies agent-level reasoning diversity while
    retaining a broadcast-style peer-response exchange. Following the original DMAD setup, we use four task-appropriate reasoning strategies: direct input--output (IO) prompting, Chain-of-Thought~\citep{wei2022chain}, Step-Back Prompting~\citep{zheng2024take}, and Self-Refine~\citep{madaan2023self,kim2023language}.

    \item \textbf{\textsc{TrueMAD}}.
    Our proposed protocol augments vanilla MAD with a one-to-one, free-form
    advice stage between response broadcasting and answer revision. For each
    ordered pair of agents, the source agent generates advice grounded in
    both its own response and the target agent's response; the advice may
    support a valid argument, challenge a specific error, or suggest a
    refinement. Each target agent then revises its solution using the
    broadcast peer responses together with the advice explicitly addressed
    to it. All other components of vanilla MAD remain unchanged, making
    \textsc{TrueMAD} and MAD a direct matched comparison.

    \item \textbf{\textsc{TrueDMAD}}.
    This variant applies the same targeted advice stage to DMAD while
    preserving its strategy-diverse agent-level reasoning conditions.
    Comparing \textsc{TrueDMAD} with DMAD evaluates whether targeted
    peer interaction remains effective when combined with reasoning
    diversity, and whether the proposed advice mechanism functions as a plug-in rather than being specific to vanilla MAD.
\end{itemize}

The primary matched comparisons are therefore \textsc{TrueMAD} versus MAD
and \textsc{TrueDMAD} versus DMAD, while promptMAD controls for the effect
of strengthening the generic debate prompt without introducing directed
agent-to-agent communication.

\section{Additional Results}
\label{section-appendix-experiment-result}

\subsection{Detailed Debate Stage Round-Wise Results}
\label{section-appendix-experiment-result-round-wise-results}

Tables~\ref{tab:appendix-qwen3-30b-aime-rounds}--\ref{tab:appendix-qwen3-30b-truthfulqa-rounds}
report the complete round-wise agent-level accuracy of \texttt{Qwen3-30B-A3B} from
R2 to R5. Each entry is the mean and sample standard deviation over three
independent seeds. 

\begin{table}[H]
    \centering
    \caption{Round-wise agent-level accuracy (\%) of \texttt{Qwen3-30B-A3B} on
    AIME. The highest mean accuracy in each round is shown in bold.}
    \label{tab:appendix-qwen3-30b-aime-rounds}
    \scriptsize
    \begin{tabular*}{\columnwidth}{@{\extracolsep{\fill}}lcccc@{}}
    \hline
    \textbf{Method} & \textbf{R2} & \textbf{R3} & \textbf{R4} & \textbf{R5} \\
    \hline
    MAD
    & $\acc{27.69}{1.70}$ & $\acc{29.35}{2.16}$
    & $\acc{30.00}{2.17}$ & $\acc{29.81}{2.49}$ \\
    promptMAD
    & $\acc{28.52}{1.37}$ & $\acc{29.81}{1.79}$
    & $\acc{30.09}{1.95}$ & $\acc{30.37}{2.39}$ \\
    DMAD
    & $\acc{29.07}{2.23}$ & $\acc{29.91}{1.25}$
    & $\acc{30.83}{1.55}$ & $\acc{30.28}{1.00}$ \\
    \textbf{\textsc{TrueMAD}}
    & $\accbold{30.74}{1.43}$ & $\accbold{34.17}{2.78}$
    & $\accbold{35.00}{2.93}$ & $\accbold{35.37}{2.36}$ \\
    \textbf{\textsc{TrueDMAD}}
    & $\acc{28.98}{0.42}$ & $\acc{31.48}{0.58}$
    & $\acc{33.52}{0.64}$ & $\acc{34.63}{1.12}$ \\
    \hline
    \end{tabular*}
\end{table}

\begin{table}[H]
    \centering
    \caption{Round-wise agent-level accuracy (\%) of \texttt{Qwen3-30B-A3B} on
    GPQA-Diamond. The highest mean accuracy in each round is shown in bold.}
    \label{tab:appendix-qwen3-30b-gpqa-rounds}
    \scriptsize
    \begin{tabular*}{\columnwidth}{@{\extracolsep{\fill}}lcccc@{}}
    \hline
    \textbf{Method} & \textbf{R2} & \textbf{R3} & \textbf{R4} & \textbf{R5} \\
    \hline
    MAD
    & $\acc{53.16}{1.46}$ & $\acc{53.79}{1.16}$
    & $\acc{54.00}{0.89}$ & $\acc{54.12}{1.29}$ \\
    promptMAD
    & $\acc{54.59}{1.56}$ & $\acc{55.85}{1.39}$
    & $\acc{55.64}{1.51}$ & $\acc{55.26}{1.52}$ \\
    DMAD
    & $\acc{54.59}{0.41}$ & $\acc{56.40}{0.60}$
    & $\acc{55.98}{0.84}$ & $\acc{56.14}{1.53}$ \\
    \textbf{\textsc{TrueMAD}}
    & $\accbold{55.30}{1.46}$ & $\acc{56.82}{0.55}$
    & $\acc{57.70}{1.70}$ & $\acc{58.00}{0.60}$ \\
    \textbf{\textsc{TrueDMAD}}
    & $\acc{55.01}{0.64}$ & $\accbold{58.50}{2.22}$
    & $\accbold{60.06}{2.44}$ & $\accbold{62.25}{1.79}$ \\
    \hline
    \end{tabular*}
\end{table}

\begin{table}[H]
    \centering
    \caption{Round-wise agent-level accuracy (\%) of \texttt{Qwen3-30B-A3B} on TruthfulQA. The highest mean accuracy in each round is shown in bold.}
    \label{tab:appendix-qwen3-30b-truthfulqa-rounds}
    \scriptsize
    \begin{tabular*}{\columnwidth}{@{\extracolsep{\fill}}lcccc@{}}
    \hline
    \textbf{Method} & \textbf{R2} & \textbf{R3} & \textbf{R4} & \textbf{R5} \\
    \hline
    MAD
    & $\acc{77.54}{0.26}$ & $\acc{77.79}{0.31}$
    & $\acc{77.83}{0.36}$ & $\acc{77.88}{0.45}$ \\
    promptMAD
    & $\acc{78.00}{0.22}$ & $\acc{77.96}{1.16}$
    & $\acc{77.96}{0.69}$ & $\acc{77.96}{0.29}$ \\
    DMAD
    & $\acc{77.75}{0.45}$ & $\acc{78.29}{0.95}$
    & $\acc{78.29}{1.91}$ & $\acc{77.79}{1.25}$ \\
    \textbf{\textsc{TrueMAD}}
    & $\acc{78.62}{0.75}$ & $\acc{79.33}{0.58}$
    & $\acc{79.46}{0.51}$ & $\acc{79.29}{0.62}$ \\
    \textbf{\textsc{TrueDMAD}}
    & $\accbold{79.00}{1.00}$ & $\accbold{80.29}{1.12}$
    & $\accbold{80.71}{0.19}$ & $\accbold{81.50}{0.38}$ \\
    \hline
    \end{tabular*}
\end{table}

\subsection{Results by Initial Correctness Composition}
\label{section-appendix-experiment-result-r1-correctness-composition}

We further group questions according to the number of agents that answer
correctly in the first round. Let $k\in\{0,1,2,3,4\}$ denote the
number of correct agents in R1. We report R5 agent-level accuracy for questions
with a wrong majority ($k\leq 1$), an even split ($k=2$), a correct majority
($k\geq 3$), and difficult cases with at least half the agents wrong ($k\leq2$), the subset shown in Figure~\ref{fig:qwen3-30b-acc-panel}(b). The first
three groups are mutually exclusive, whereas the last group is the union of
the wrong-majority and even-split groups used in the main-text analysis. Each
value is a micro-average over all agent--question pairs and three seeds within
the corresponding R1 group.

\begin{table}[H]
    \centering
    \caption{R5 agent-level accuracy (\%) of \texttt{Qwen3-30B-A3B} on AIME,
    stratified by the R1 correctness composition. The highest accuracy in each column is shown in bold.}
    \label{tab:appendix-qwen3-30b-aime-r1-groups}
    \scriptsize
    \begin{tabular*}{\columnwidth}{@{\extracolsep{\fill}}lcccc@{}}
    \hline
    \textbf{Method}
    & \textbf{Wrong maj.} & \textbf{Split}
    & \textbf{Correct maj.} & \textbf{$\geq$ Half wrong} \\
    & $k\leq 1$ & $k=2$ & $k\geq 3$ & $k\leq 2$ \\
    \hline
    MAD
    & $7.6$ & $\mathbf{93.4}$ & $98.4$ & $15.0$ \\
    promptMAD
    & $8.1$ & $92.1$ & $\mathbf{100.0}$ & $15.3$ \\
    DMAD
    & $9.1$ & $86.8$ & $97.4$ & $15.8$ \\
    \textbf{\textsc{TrueMAD}}
    & $\mathbf{14.7}$ & $\mathbf{93.4}$ & $\mathbf{100.0}$
    & $\mathbf{21.4}$ \\
    \textbf{\textsc{TrueDMAD}}
    & $14.4$ & $89.5$ & $98.4$ & $20.8$ \\
    \hline
    \end{tabular*}
\end{table}

\begin{table}[H]
    \centering
    \caption{R5 agent-level accuracy (\%) of \texttt{Qwen3-30B-A3B} on
    GPQA-Diamond, stratified by the R1 correctness composition. The highest accuracy in each column is shown in bold.}
    \label{tab:appendix-qwen3-30b-gpqa-r1-groups}
    \scriptsize
    \begin{tabular*}{\columnwidth}{@{\extracolsep{\fill}}lcccc@{}}
    \hline
    \textbf{Method}
    & \textbf{Wrong maj.} & \textbf{Split}
    & \textbf{Correct maj.} & \textbf{$\geq$ Half wrong} \\
    & $k\leq 1$ & $k=2$ & $k\geq 3$ & $k\leq 2$ \\
    \hline
    MAD
    & $9.8$ & $54.7$ & $93.7$ & $20.9$ \\
    promptMAD
    & $10.9$ & $55.0$ & $95.1$ & $21.8$ \\
    DMAD
    & $11.9$ & $54.7$ & $\mathbf{96.2}$ & $22.5$ \\
    \textbf{\textsc{TrueMAD}}
    & $17.2$ & $62.8$ & $93.2$ & $28.5$ \\
    \textbf{\textsc{TrueDMAD}}
    & $\mathbf{21.0}$ & $\mathbf{72.8}$ & $96.1$
    & $\mathbf{33.8}$ \\
    \hline
    \end{tabular*}
\end{table}

\begin{table}[H]
    \centering
    \caption{R5 agent-level accuracy (\%) of \texttt{Qwen3-30B-A3B} on
    TruthfulQA, stratified by the R1 correctness composition. The highest accuracy in each column is shown in bold.}
    \label{tab:appendix-qwen3-30b-truthfulqa-r1-groups}
    \scriptsize
    \begin{tabular*}{\columnwidth}{@{\extracolsep{\fill}}lcccc@{}}
    \hline
    \textbf{Method}
    & \textbf{Wrong maj.} & \textbf{Split}
    & \textbf{Correct maj.} & \textbf{$\geq$ Half wrong} \\
    & $k\leq 1$ & $k=2$ & $k\geq 3$ & $k\leq 2$ \\
    \hline
    MAD
    & $6.6$ & $59.0$ & $\mathbf{97.3}$ & $15.7$ \\
    promptMAD
    & $11.4$ & $57.0$ & $96.3$ & $19.4$ \\
    DMAD
    & $13.8$ & $63.0$ & $95.1$ & $22.4$ \\
    \textbf{\textsc{TrueMAD}}
    & $15.7$ & $58.0$ & $96.9$ & $23.1$ \\
    \textbf{\textsc{TrueDMAD}}
    & $\mathbf{23.3}$ & $\mathbf{73.0}$ & $97.0$
    & $\mathbf{32.0}$ \\
    \hline
    \end{tabular*}
\end{table}

\subsection{Detailed Process-Level Scores}
\label{section-appendix-experiment-result-detailed-process-level-scores}

Tables~\ref{tab:appendix-qwen3-30b-peer-quality} and~\ref{tab:appendix-qwen3-30b-process-metrics} report the numerical values
visualized in Figure~\ref{fig:process-level-metrics}. The first table gives
the three LLM-judge dimensions of peer-deliberation quality: Rebuttal
Precision (RP), Evidence Independence (EI), and Novel Contribution (NC). The
second table gives Correction Preference Lift (CPL), Asymmetric Majority
Escape (AME), and Correct-Minority Propagation (CMP$_{1v3}$). AME is reported
in percentage points and CMP$_{1v3}$ in percent; higher values are better for
all six metrics.

\begin{table}[H]
    \centering
    \caption{Peer-deliberation quality scores of \texttt{Qwen3-30B-A3B} corresponding
    to Figure~\ref{fig:process-level-metrics}(a). RP, EI, and NC denote
    Rebuttal Precision, Evidence Independence, and Novel Contribution,
    respectively. The highest score in each column is shown in bold.}
    \label{tab:appendix-qwen3-30b-peer-quality}
    \scriptsize
    \begin{tabular*}{\textwidth}{@{\extracolsep{\fill}}lccccccccc@{}}
    \hline
    \multirow{2}{*}{\textbf{Method}}
    & \multicolumn{3}{c}{\textbf{AIME}}
    & \multicolumn{3}{c}{\textbf{GPQA-Diamond}}
    & \multicolumn{3}{c}{\textbf{TruthfulQA}} \\
    \cline{2-4}\cline{5-7}\cline{8-10}
    & \textbf{RP} & \textbf{EI} & \textbf{NC}
    & \textbf{RP} & \textbf{EI} & \textbf{NC}
    & \textbf{RP} & \textbf{EI} & \textbf{NC} \\
    \hline
    MAD
    & $3.16$ & $4.04$ & $2.76$
    & $3.86$ & $4.83$ & $3.36$
    & $3.84$ & $4.49$ & $3.11$ \\
    promptMAD
    & $3.41$ & $4.42$ & $\mathbf{3.13}$
    & $4.53$ & $5.63$ & $\mathbf{4.36}$
    & $4.99$ & $6.12$ & $\mathbf{5.17}$ \\
    DMAD
    & $2.88$ & $4.45$ & $2.81$
    & $4.23$ & $5.96$ & $3.85$
    & $5.13$ & $6.39$ & $4.38$ \\
    \textbf{\textsc{TrueMAD}}
    & $\mathbf{3.82}$ & $\mathbf{5.03}$ & $3.12$
    & $\mathbf{4.86}$ & $6.27$ & $4.23$
    & $\mathbf{5.42}$ & $\mathbf{6.93}$ & $5.12$ \\
    \textbf{\textsc{TrueDMAD}}
    & $3.51$ & $4.88$ & $3.08$
    & $4.68$ & $\mathbf{6.30}$ & $4.18$
    & $5.24$ & $6.75$ & $4.80$ \\
    \hline
    \end{tabular*}
\end{table}

\begin{table}[H]
    \centering
    \caption{Correction-preference and consensus-robustness scores of
    \texttt{Qwen3-30B-A3B} corresponding to
    Figure~\ref{fig:process-level-metrics}(b). CPL is unitless, AME is
    reported in percentage points, and CMP$_{1v3}$ is reported in percent.
    The highest score in each column is shown in bold.}
    \label{tab:appendix-qwen3-30b-process-metrics}
    \scriptsize
    \begin{tabular*}{\textwidth}{@{\extracolsep{\fill}}lccccccccc@{}}
    \hline
    \multirow{2}{*}{\textbf{Method}}
    & \multicolumn{3}{c}{\textbf{AIME}}
    & \multicolumn{3}{c}{\textbf{GPQA-Diamond}}
    & \multicolumn{3}{c}{\textbf{TruthfulQA}} \\
    \cline{2-4}\cline{5-7}\cline{8-10}
    & \textbf{CPL} & \textbf{AME} & \textbf{CMP$_{1v3}$}
    & \textbf{CPL} & \textbf{AME} & \textbf{CMP$_{1v3}$}
    & \textbf{CPL} & \textbf{AME} & \textbf{CMP$_{1v3}$} \\
    \hline
    MAD
    & $0.85$ & $4.6$ & $25.0$
    & $1.28$ & $3.7$ & $25.4$
    & $4.31$ & $2.1$ & $14.1$ \\
    promptMAD
    & $0.89$ & $7.2$ & $33.3$
    & $1.42$ & $4.9$ & $27.7$
    & $3.92$ & $4.6$ & $34.3$ \\
    DMAD
    & $0.80$ & $6.8$ & $40.4$
    & $1.43$ & $10.1$ & $36.0$
    & $3.71$ & $10.9$ & $47.4$ \\
    \textbf{\textsc{TrueMAD}}
    & $\mathbf{1.96}$ & $7.1$ & $\mathbf{55.0}$
    & $1.65$ & $6.6$ & $32.4$
    & $4.81$ & $8.8$ & $25.0$ \\
    \textbf{\textsc{TrueDMAD}}
    & $1.01$ & $\mathbf{10.9}$ & $47.9$
    & $\mathbf{2.00}$ & $\mathbf{12.8}$ & $\mathbf{47.3}$
    & $\mathbf{5.49}$ & $\mathbf{13.7}$ & $\mathbf{54.8}$ \\
    \hline
    \end{tabular*}
\end{table}

\section{Prompt Templates}
\label{section-appendix-prompt-templates}

This appendix reports the exact prompt templates used in our experiments.
Expressions enclosed in braces are runtime placeholders and are replaced with
the corresponding question, response, or intermediate output. Unless stated
otherwise, the experiments use four agents. Text inside the shaded boxes is
reproduced verbatim: capitalization, wording, spelling, line breaks, and
spacing are preserved to document the implemented protocol rather than a
post-hoc edited version.

Each method executes its own initialization. MAD, promptMAD, and
\textsc{TrueMAD} use the homogeneous initial prompt in
Section~\ref{app:mad-initial-prompt}, whereas DMAD and \textsc{TrueDMAD} use
the four role-specific initializers in
Section~\ref{app:dmad-initial-prompts}. The advice template in
Section~\ref{app:truemad-advice-prompt} is used only to generate targeted
peer suggestions. After receiving the previous-round answers and suggestions,
each recipient applies the same revision procedure as its underlying method.

\subsection{Runtime Placeholders and Output Formats}
\label{app:prompt-runtime-fields}

Table~\ref{tab:prompt-placeholders} defines the placeholders appearing in the
templates.

\begin{table}[H]
    \centering
    \caption{Runtime placeholders used in the prompt templates.}
    \label{tab:prompt-placeholders}
    \small
    \setlength{\tabcolsep}{5pt}
    \begin{tabular}{@{}p{0.25\textwidth}p{0.69\textwidth}@{}}
        \toprule
        \textbf{Placeholder} & \textbf{Runtime value} \\
        \midrule
        \texttt{\{question\}} &
        Question text; for multiple-choice datasets, this already includes
        the answer options. \\
        \texttt{\{format\_instruction\}} &
        Dataset-specific output instruction shown below. \\
        \texttt{\{debater\_id\}} &
        Integer identifier of the current debater, from \texttt{0} to
        \texttt{3}. \\
        \texttt{\{answer\_i\}} &
        Previous-round answer of debater \texttt{i}, including the current
        debater's own answer. \\
        \texttt{\{own\_answer\}} &
        Previous-round answer of the agent producing advice. \\
        \texttt{\{target\_id\}} &
        Integer identifier of the debater receiving advice. \\
        \texttt{\{target\_answer\}} &
        Previous-round answer of the target debater. \\
        \texttt{\{advice\}} &
        Concatenated targeted suggestions addressed to the current agent. \\
        \texttt{\{self\_answer\}} &
        Current DMAD agent's previous-round final answer. \\
        \texttt{\{label\}} &
        DMAD agent label: \texttt{A}, \texttt{B}, \texttt{C}, or \texttt{D}. \\
        \texttt{\{other\_answer\}} &
        Previous-round final answer of another DMAD agent. \\
        \texttt{\{step1\}}, \texttt{\{step2\}} &
        Text returned by a previous step within the same turn. \\
        \bottomrule
    \end{tabular}
\end{table}

\paragraph{Dataset-specific output instructions.}
The following string is substituted for \texttt{\{format\_instruction\}}
according to the answer format of the dataset.

\promptusage{Used for open-ended AIME items.}
\begin{promptbox}{AIME output format}
Explain your answer, putting the choice in the form "\boxed{answer}" at the end of your response
\end{promptbox}

\promptusage{Used for multiple-choice GPQA-Diamond and TruthfulQA items.}
\begin{promptbox}{Multiple-choice output format}
Explain your answer, putting the choice in the form "\boxed{choice letter}" (For example:\boxed{A} or \boxed{B} or \boxed{C} or \boxed{D}) at the end of your response
\end{promptbox}

\subsection{Initial-Stage Prompts}
\label{app:initial-stage-prompts}

\promptsubsubsection{MAD-Family Initial Prompt}
\label{app:mad-initial-prompt}

\promptusage{Used independently by MAD, promptMAD, and \textsc{TrueMAD}.}
\begin{promptbox}{MAD-family initial prompt}
{question}
{format_instruction}
\end{promptbox}

\promptsubsubsection{DMAD Initial Prompts}
\label{app:dmad-initial-prompts}

DMAD and \textsc{TrueDMAD} employ four distinct reasoning manners. The
following templates are used for their respective initializations.

\promptagent{Reasoning Manner 1 (IO).}
\promptusage{One-step direct generation.}
\begin{promptbox}{DMAD initial prompt -- Reasoning Manner 1}
{question}

{format_instruction}
\end{promptbox}

\promptagent{Reasoning Manner 2 (CoT).}
\promptusage{One-step chain-of-thought generation.}
\begin{promptbox}{DMAD initial prompt -- Reasoning Manner 2}
{question}

Let's think step by step.

{format_instruction}
\end{promptbox}

\promptagent{Reasoning Manner 3 (SBP).}
\promptusage{Two-step generation; the second call receives the output of the first through \texttt{\{step1\}}.}
\begin{promptbox}{DMAD initial prompt -- Reasoning Manner 3, Step 1}
{question}

You are an expert and your task is to extract the concepts and principles involved in solving the problem

{format_instruction}
\end{promptbox}

\begin{promptbox}{DMAD initial prompt -- Reasoning Manner 3, Step 2}
{question}

Concepts and principles in solving this problem:
{step1}

Solve the problem step by step according to the concepts and principles

{format_instruction}
\end{promptbox}

\promptagent{Reasoning Manner 4 (SR).}
\promptusage{Three-step self-refinement; later calls receive earlier outputs through \texttt{\{step1\}} and \texttt{\{step2\}}.}
\begin{promptbox}{DMAD initial prompt -- Reasoning Manner 4, Step 1}
{question}

Let's think step by step.

{format_instruction}
\end{promptbox}

\begin{promptbox}{DMAD initial prompt -- Reasoning Manner 4, Step 2}
{question}

{format_instruction}

Your previous answer:
{step1}

Review your previous answer and find problems with your answers.
\end{promptbox}

\begin{promptbox}{DMAD initial prompt -- Reasoning Manner 4, Step 3}
{question}

Your previous answer:
{step1}

Problems in your previous answer:
{step2}

Based on the problems of your previous answer, improve your answer

{format_instruction}
\end{promptbox}

\subsection{Broadcast Debate Prompts}
\label{app:broadcast-debate-prompts}

MAD and promptMAD retain only the immediately preceding round. 

\promptsubsubsection{MAD}

\promptusage{Used at every debate-stage turn of vanilla MAD.}
\begin{promptbox}{MAD debate prompt}
{question}
{format_instruction}

Your previous answer:
{self_answer}

Other debaters' answers:
Agent {label}:
{other_answer}

You are Debater {debater_id}. Given the other debaters' answers, update your answer.
\end{promptbox}

\promptsubsubsection{promptMAD}

\promptusage{Uses the same debate context as MAD, with a more explicit engagement instruction.}
\begin{promptbox}{promptMAD debate prompt}
{question}
{format_instruction}

Your previous answer:
{self_answer}

Other debaters' answers:
Agent {label}:
{other_answer}

You are participating in a debate about answering questions. Your role is the debater {debater_id}. Given the answers from other debaters, evaluate their answers based on your own understanding and provide support, counterarguments, or further questions. Update your answer if needed. {format_instruction}
\end{promptbox}

\subsection{DMAD Debate Prompts}
\label{app:dmad-debate-prompts}

At each debate turn, a DMAD agent receives the question, its own previous
final answer, and the other agents' previous final answers. Intermediate
outputs generated by other agents are not included. The header below appears
once, followed by one \texttt{Agent \{label\}} entry for every agent other
than the current one.

\begin{promptbox}{DMAD other-answer block}
Other debaters' answers:
Agent {label}:
{other_answer}
\end{promptbox}

\promptsubsubsection{Reasoning Manner 1 (IO)}

\promptusage{One-step debate update.}
\begin{promptbox}{DMAD debate prompt -- Reasoning Manner 1}
{question}

{format_instruction}

Your previous answer:
{self_answer}

Other debaters' answers:
Agent {label}:
{other_answer}

Given the answers from other debaters, update your answer.

{format_instruction}
\end{promptbox}

\promptsubsubsection{Reasoning Manner 2 (CoT)}

\promptusage{One-step debate update with an explicit step-by-step instruction.}
\begin{promptbox}{DMAD debate prompt -- Reasoning Manner 2}
{question}

{format_instruction}

Your previous answer:
{self_answer}

Other debaters' answers:
Agent {label}:
{other_answer}

Given the answers from other debaters, update your answer.

Let's think step by step.

{format_instruction}
\end{promptbox}

\promptsubsubsection{Reasoning Manner 3 (SBP)}

\promptusage{Two-step debate update; the second step receives the extracted concepts and principles through \texttt{\{step1\}}.}
\begin{promptbox}{DMAD debate prompt -- Reasoning Manner 3, Step 1}
{question}

Your previous answer:
{self_answer}

Other debaters' answers:
Agent {label}:
{other_answer}

Given the answers from other debaters, you are an expert and your task is to extract the concepts and principles involved in updating your answers

{format_instruction}
\end{promptbox}

\begin{promptbox}{DMAD debate prompt -- Reasoning Manner 3, Step 2}
{question}

Your previous answer:
{self_answer}

Other debaters' answers:
Agent {label}:
{other_answer}

Concepts and principles in updating your answer:
{step1}

Given the answers from other debaters, update your answer step by step according to the concepts and principles.

{format_instruction}
\end{promptbox}

\promptsubsubsection{Reasoning Manner 4 (SR)}

\promptusage{Three-step debate update; the second and third steps review and improve the response generated in the first step.}
\begin{promptbox}{DMAD debate prompt -- Reasoning Manner 4, Step 1}
{question}

Your previous answer:
{self_answer}

Other debaters' answers:
Agent {label}:
{other_answer}

Given the answers from other debaters, update your answer.

Let's think step by step.

{format_instruction}
\end{promptbox}

\begin{promptbox}{DMAD debate prompt -- Reasoning Manner 4, Step 2}
{question}

{format_instruction}

Your previous answer:
{self_answer}

Other debaters' answers:
Agent {label}:
{other_answer}

Given the answers from other debaters, your task is updating your answer.

Your previous answer:
{step1}

Review your previous answer and find problems with your answers.
\end{promptbox}

\begin{promptbox}{DMAD debate prompt -- Reasoning Manner 4, Step 3}
{question}

Your previous answer:
{self_answer}

Other debaters' answers:
Agent {label}:
{other_answer}

Given the answers from other debaters, your task is updating your answer.

Your previous answer:
{step1}

Problems in your previous answer:
{step2}

Based on the problems of your previous answer, improve your answer

{format_instruction}
\end{promptbox}

\Needspace{18\baselineskip}
\subsection{\textsc{TrueMAD} Targeted-Advice Prompt}
\label{app:truemad-advice-prompt}

Before each debate-stage revision, every other agent independently comments on
the current target agent, resulting in one call for each ordered
source--target pair. The model input consists of the question and output
instruction, the source agent's previous answer, the target's previous answer, and
the evaluation instruction below. In the implemented string, there is no
space after \texttt{answer:}.

\promptusage{Used by the \textsc{TrueMAD} interaction plug-in before the target agent performs its debate-stage revision.}
\begin{promptbox}{\textsc{TrueMAD} targeted-advice prompt}
{question}
{format_instruction}
Your answer in the previous round:
{own_answer}
Debater {target_id}'s answer:{target_answer}
Evaluate Debater {target_id}'s answer based on your own understanding and provide support, counterarguments, or further questions.
\end{promptbox}

The generated suggestions are concatenated by source-agent identifier and
inserted into the target's normal debate context using the following block:
\begin{promptbox}{Targeted-advice context block}
Targeted peer advice addressed to you:
{advice}
\end{promptbox}
For one-step MAD-family updaters, this block appears after the previous-round
answers and before the closing revision instruction. The answer block still
contains all four agents' previous-round answers, including the target
agent's own answer. For multi-step DMAD updaters, the advice block is included
in the first step; later steps receive the resulting intermediate output
exactly as in the corresponding DMAD procedure. The target therefore uses
its underlying updater without modification: MAD and promptMAD retain their
one-step revision procedures, while DMAD and \textsc{TrueDMAD} retain the
corresponding IO, chain-of-thought, Step-Back, and Self-Refine procedures
shown above.

\subsection{LLM-as-a-Judge Prompt for Level 1}
\label{app:level1-judge-prompt}

For the Level 1 utterance-level peer deliberation evaluation, an LLM judge receives the
debate history and the model's current reply, and independently scores
\emph{Rebuttal Precision}, \emph{Evidence Independence}, and
\emph{Novel Contribution}. Each dimension is assigned an integer score from
1 to 10. The rubric explicitly separates debate-process quality from answer
correctness and requires a machine-readable JSON response. The following
template reproduces the judge instruction used in our evaluation.

\promptusage{Used to score each evaluated debate-stage reply for the three Level 1 process-quality dimensions.}
\begin{promptbox}{Level 1 LLM-as-a-Judge prompt}
You are an expert debate evaluator. Given the debater history and the model's current reply, score it from 1-10 on three dimensions. 

- Rebuttal Precision: Does it address a specific argument or critique, locate the exact error, explain why it is wrong, and show its effect? Generic agreement, disagreement, or summary scores low.

- Evidence Independence: Does it change or maintain its position based on reasoning evidence rather than consensus or imitation? A change should identify the corrected error; maintaining a position should specifically rebut the critique. Independent reasoning scores high; copying or following the majority scores low.

- Novel Contribution: Does it add a relevant new idea, reasoning step, alternative approach, counterexample, missing condition, or concrete check? Repetition and paraphrasing score low.

Evaluate reasoning rather than answer agreement or confidence. Score only what the current statement demonstrates in context. Use integers only.

Respond strictly as JSON:

{
  "rebuttal_precision": <int>,
  "evidence_independence": <int>,
  "novel_contribution": <int>
}
\end{promptbox}

\Needspace{24\baselineskip}
\subsection{Auxiliary-Call Control Prompts}
\label{app:auxiliary-call-control-prompts}

The following prompts implement the two auxiliary-call controls used in
RQ4. Here, \texttt{\{format\_suffix\}} denotes the dataset-specific output
instruction, and \texttt{\{self\_label\}} identifies the current debater.

\promptsubsubsection{ComputeMAD}

Each call receives all four previous-round answers without specifying a
comparison target.
\promptusage{Invoked 12 times before each revision in R2--R5 to generate non-targeted debate notes.}
\begin{promptbox}{ComputeMAD note-generation prompt}
{question}
{format_suffix}
Available answers:
Debater 0:
{answer_0}
Debater 1:
{answer_1}
Debater 2:
{answer_2}
Debater 3:
{answer_3}
Examine the available answers and produce one useful debate note, such as a correction, verification, or missing consideration.
\end{promptbox}

The 12 calls produce one note each. The notes are then randomly assigned
without replacement across the four agents, with each agent receiving three
notes as advice for its next-round revision.

\promptsubsubsection{Self-AdviceMAD}

Each call receives only the current agent's previous-round answer and requests
a critique rather than a revised solution.
\promptusage{Invoked independently three times per agent before each revision in R2--R5.}
\begin{promptbox}{Self-AdviceMAD note-generation prompt}
{question}
{format_suffix}
Your role is Debater {self_label}.
Your answer in the previous round:
{self_answer}
Critique your own previous answer and produce one useful note, such as a correction, verification, or missing consideration. Do not write a full revised solution; write a critique only.
\end{promptbox}

For each agent, the three independently generated notes are directly used as
advice for that agent's next-round revision.

\end{document}